\documentclass[runningheads]{llncs}
\usepackage{graphicx}
\usepackage{multirow}
\usepackage{lscape}
\begin{document}
\title{CoNTACT: \\A Dutch COVID-19 Adapted BERT for Vaccine Hesitancy and Argumentation Detection\thanks{This research received funding from the Vaccine Confidence Fund (VCF) \url{https://vaccineconfidencefund.org/}}}
\titlerunning{CoNTACT}
\author{Jens Lemmens \and
Jens Van Nooten \and
Tim Kreutz \and
Walter Daelemans}
\authorrunning{Lemmens et al.}
\institute{CLiPS (University of Antwerp)\\Lange Winkelstraat 40, 2000 Antwerp\\
\email{firstname.lastname@uantwerpen.be}}
\maketitle              % typeset the header of the contribution
\begin{abstract}
We present CoNTACT\footnote{The model is available at \url{https://huggingface.co/clips/contact}}: a Dutch language model adapted to the domain of COVID-19 tweets. The model was developed by continuing the pre-training phase of RobBERT \cite{delobelle:2020} by using 2.8M Dutch COVID-19 related tweets posted in 2021. In order to test the performance of the model and compare it to RobBERT, the two models were tested on two tasks: (1) binary vaccine hesitancy detection and (2) detection of arguments for vaccine hesitancy. For both tasks, not only Twitter but also Facebook data was used to show cross-genre performance. In our experiments, CoNTACT showed statistically significant gains over RobBERT in all experiments for task 1. For task 2, we observed substantial improvements in virtually all classes in all experiments. An error analysis indicated that the domain adaptation yielded better representations of domain-specific terminology, causing CoNTACT to make more accurate classification decisions. 

\end{abstract}
\section{Introduction} \label{intro}
We present CoNTACT (Contextual Neural Transformer Adapted to COVID-19 Tweets). The model was developed by fine-tuning RobBERT \cite{delobelle:2020} (a RoBERTa-base model \cite{liu:2019} pre-trained on Dutch data) on masked language modeling using 2.8M Dutch-language tweets related to COVID-19 that were posted in 2021. The model was evaluated on two tasks: (1) binary vaccine hesitancy classification and (2) classification of arguments for vaccine hesitancy. In order to measure the effect of the domain adaptation, the results were compared to out-of-the-box RobBERT. Moreover, the aforementioned tasks were not only performed on tweets, but also on Facebook comments to show the cross-genre benefits of the domain adaptation. Afterwards, a qualitative error analysis was conducted to show where CoNTACT improved compared to RobBERT and where it could potentially improve further.
In earlier research, an English language model pre-trained on COVID-19 related tweets (COVID-Twitter-BERT) was developed \cite{muller:2020}. We apply the same methodology for the first time to Dutch and extensively test the effect on two COVID-19 related classification tasks. 

\section{Related research} \label{research}

Traditional machine learning assumes that models are trained and tested on large amounts of data from the same domain, which is not always feasible due to lack of labelled data. Transfer learning, which involves the transfer of knowledge from one domain to another, is a technique that has been utilized successfully in machine learning, both in NLP (e.g. \cite{weiss:2016}, \cite{durrani:2021}, \cite{rostami:2021}) and computer vision (e.g. \cite{wang:2018}, \cite{voulodimos:2018}, \cite{xu:2019}) to combat this issue. An effective approach to transfer learning used frequently in recent years is the pre-training of language models, such as BERT \cite{devlin:2018}, on large amounts of unsupervised data. The knowledge from this pre-training phase is then transferred to the subsequent fine-tuning phase on task- and domain-specific data, which has shown significant improvements on several benchmark datasets, e.g., \cite{leong:2020} and \cite{basile:2019}. Subsequently, several language- and domain-specific adaptations of language models have been developed for non-English data or to further improve the performance of the original models on specific tasks. Examples are BERTje and RobBERT \cite{devries:2019} \cite{delobelle:2020}, the Dutch equivalents of BERT and RoBERTa, respectively), and CamemBERT \cite{martin:2019}, a French BERT model.

Domain adaptation, a special case of transfer learning where the model is first trained on unsupervised data from the domain of an intended task, aims to improve results even further ``by minimizing the difference between the domain distributions" (\cite{farahani:2020}, p. 1), thus creating a model that optimally learns from the training data. Regarding the domain of COVID-19, COVID-Twitter-BERT \cite{muller:2020}, a BERT-large model pre-trained on COVID-19 tweets, has shown statistically significant gains over the baseline BERT-large in various applications, including vaccine stance classification.

In other research related to vaccine stance classification, various rule-based, statistical and deep learning approaches for the classification of stance towards vaccines have been compared \cite{joshi:2018}. Concretely, the task consisted of multiclass classification of vaccine stance in social media messages (``for", ``againts" or ``undecided"). The authors concluded that both pre-trained language models and statistical ensemble models achieved equally high results on this task. This work focused on vaccine stance in general, but since the start of the COVID-19 pandemic vaccine stance classification has become almost inextricably linked to COVID-19 due to its societal relevance. An example is \cite{weinzierl:2021} who present CoVaxLies, a COVID-19 vaccine misinformation dataset, and demonstrate that their model, based on knowledge graphs, outperforms widely used classification methods for the detection of vaccine misinformation, an important cause of vaccine hesitancy. 

Specifically for Dutch, \cite{wang:2020} collected Dutch tweets using keywords, and comments from Reddit and Nu.nl\footnote{Nu.nl is a Dutch news website that allows visitors to comment on news articles} threads related to COVID-19 in order to investigate polarity (``positive"/``negative") and stance (``support"/``reject"/``other") towards face masks and the social distance measure between March and October 2020. For polarity analysis, the Pattern library \cite{desmedt:2012} was used, whereas manual annotations were used to train a stance classifier consisting of a linear feed forward neural network using stochastic gradient descend and a subword embedding layer, which achieved a test set accuracy of 65\%. After applying the polarity analyzer and stance classifier to the above-mentioned data, it was shown that a more negative polarity was found in COVID-19 related messages than in a subset of messages that were unrelated to COVID-19. More specifically, a more negative polarity (and also stance) was found in messages mentioning face masks than in messages mentioning the social distancing measure. The various social media platforms that were used showed similar trends over time.

\section{Methodology}
\label{methodology}
\subsection{Domain adaptation}
For the development of CoNTACT, we utilize RobBERT \cite{delobelle:2020}, a Dutch RoBERTa model with 12 attention layers and 12 heads with 117M parameters trained on the Dutch segment of the OSCAR corpus (6.6B words). In line with \cite{xiaochuang:2019}, we approached adapting RobBERT by continuing its pre-training phase, that is by performing masked language modeling. For this task, we scraped Dutch-language tweets posted in 2021 using the Twitter API and the keyword method described in \cite{kreutz:2019}. Then, all tweets related to COVID-19 were filtered from this Twitter collection using regular expressions based on inflected forms, part-of-speech tag variations and spelling variations of the keywords shown in Table \ref{tab:key_words}.

\begin{table}[b]
\centering
\caption{Keyword lemmas used to construct regular expressions for collecting COVID-19 related tweets (translated from Dutch to English).}
\begin{tabular}{|c|}
\hline
\multicolumn{1}{|c|}{\textbf{Key words}} \\ \hline
corona, COVID-19, SARS-CoV-2, virologist, virus \\ \hline
\begin{tabular}[c]{@{}c@{}}vaccine, vaccinate, Astrazeneca, Pfizer, Moderna, \\ Johnson \& Johnson, Curevac, Sputnik\end{tabular} \\ \hline
\begin{tabular}[c]{@{}c@{}}mouth mask, social distancing, bubble, contact tracing, \\ quarantine, lockdown, curfew, 1.5m, cuddle contact\\ \hline \end{tabular}
\end{tabular}
\label{tab:key_words}
\end{table}

Afterwards, all duplicates and retweets were filtered from this subset of COVID-19 related tweets. To detect retweets, we based ourselves on the ``retweet status" attribute returned by the Twitter API and searched for tweets beginning with ``RT @". Finally, the FastText language detector was used to remove all tweets that were not written in Dutch \cite{joulin:2016}. In the end, 2.8M tweets (66.8M tokens, split by whitespace) remained for the domain adaptation, which were anonymized by replacing all tokens starting with ``@" by ``@USER". In order to estimate the precision, 300 randomly selected tweets were manually read and it was determined whether they were Dutch and relevant to the domain of COVID-19. This manual evaluation shows that our keyword extraction method has a precision of 90.0\%. False positives included messages about other viruses and vaccines, such as the flu/influenza, and a single tweet in Afrikaans that did not get detected by the language detector.

For the domain adaptation, the 2.8M tweets mentioned above were used to continue RobBERT's pre-training phase for 4 epochs, using the default learning rate and the largest batch size that fit working memory (32). A loss of 1.702 was achieved on a validation set consisting of 20\% of our data.

\subsection{Data and experiments}

To determine the effect of the domain adaptation, i.e., whether CoNTACT performs significantly better than RobBERT on tasks involving social media data related to COVID-19, the models were tested on two classification tasks: (1) vaccine hesitancy detection and (2) detection of arguments for vaccine hesitancy. The corpus used for the classification tasks was first described in \cite{lemmens:2022}, it consists of approx. 8,800 tweets and 5,200 Facebook comments annotated for vaccine stance and argumentation. Regarding the stance, possible class labels were ``anti-vaccination", ``vaccine-hesitant", ``neutral" and ``pro-vaccination", but these were converted to binary labels: ``anti-vaccination" and ``vaccine-hesitant" comprise the ``hesitant" category, whereas the ``not hesitant" category consists of all ``neutral" and ``pro" comments. The annotation scheme for vaccine hesitancy arguments on the other hand consisted of the following labels: 

\begin{enumerate}
    \item \textbf{Development}: messages that express worry about the development, testing  methodology, distribution and public access
    of vaccines.
    \item \textbf{Liberty}: messages that express concerns about how vaccines and vaccine laws affect civil liberty and personal freedom.
    \item \textbf{Institutional motives}: messages expressing mistrust in motives of political or economic entities involved with vaccines.
    \item \textbf{Efficacy}: messages claiming that vaccines are not efficient (enough) or unnecessary.
    \item \textbf{Safety}: messages that express worry towards the safety of the vaccines and their side effects.
    \item \textbf{Criticism on the vaccination strategy}: messages criticizing the  government's vaccination strategy/campaign.
    \item \textbf{Alternative medicine}: messages that prefer other means of protection over vaccines.
    \item \textbf{Conspiracy theories}: messages that spread conspiracy theories about vaccines.
\end{enumerate}

\clearpage
For vaccine hesitancy detection, both RobBERT and CoNTACT were fine-tuned with 10-fold cross validation. These cross validation experiments were performed in same-genre settings (fine-tuning and testing on tweets only; fine-tuning and testing on Facebook comments only) and mixed-genre settings (fine-tuning and testing on both Facebook and Twitter). Additionally, cross-genre experiments were conducted by fine-tuning on all Twitter data and testing on all Facebook data (and vice versa) in order to show the usefulness of CoNTACT when no data from an intended platform is available for fine-tuning. In order to avoid overfitting on a certain class or platform due to unbalanced data, a subset that was balanced by class and social media platform was used. The statistics of this subset can be found in Table \ref{tab:stance_data}. For all experiments, the default batch size (8) and learning rate (5e-5) was used and fine-tuning was performed for 4 epochs.

For the argumentation detection task, 8,439 tweets and 3,917 Facebook comments were used (i.e. all of the available vaccine-hesitant messages). The distribution of the arguments varies across the two social media platforms, as can be derived from Table \ref{tab:arguments_data}. Further, it should be noted that vaccine-hesitant entries without any clear argumentation were used as negative examples for the models to learn from. Similarly to the stance detection task, the aforementioned data was used to fine-tune both RobBERT and our CoNTACT model. For the same- and mixed genre experiments, cross validation was used, whereas a train-test split was used for the cross-genre experiments. Since the data is heavily unbalanced in terms of argument distribution, however, we chose to conduct experiments with 5-fold instead of 10-fold cross validation in order to preserve more entries per test set. For all experiments, the default batch size (8) and learning rate (5e-5) were used and fine-tuning was performed for 4 epochs. 

\begin{table}[t]
\centering
\caption{Vaccine hesitancy data used for the cross validation experiments.}
\begin{tabular}{|c|c|c|c|}
\hline
\textbf{Class} & \textbf{Twitter} & \textbf{Facebook} & \textbf{Total} \\ \hline
hesitant & 1250 & 1250 & 2500 \\
non-hesitant & 1250 & 1250 & 2500 \\
Total & 2500 & 2500 & 5000 \\ \hline 
\end{tabular}
\label{tab:stance_data}
\end{table}

\begin{table}[t]
\centering
\caption{Vaccine hesitancy arguments data used for the cross validation experiments.}
\begin{tabular}{|c|c|c|c|}
\hline
\textbf{Class}                                                               & \textbf{Twitter} & \textbf{Facebook} & \textbf{Total} \\ \hline
alternative medicine  & 175   & 56    & 175     \\
conspiracy theory        & 687   & 228   & 915     \\
criticism on vaccination strategy & 979              & 1,222             & 2,201         \\
development           & 565   & 511   & 1,076   \\
efficacy              & 860   & 400   & 1,260   \\
institutional motives & 1,189 & 312   & 2,131   \\
safety                & 1,493 & 1,416 & 2,909   \\
none                  & 1,153  & 298   & 1,451   \\ \hline
n messages            & 8,439 & 3,917 & 12,356 \\ \hline
\end{tabular} 
\label{tab:arguments_data}
\end{table}

\section{Results}

\subsection{Vaccine hesitancy detection}

In Table \ref{tab:stance_results}, the results of the experiments for vaccine hesitancy detection can be found. For the same-genre and mixed-genre experiments, the provided results (precision, recall, F1-score) are the averages of the test set scores on the positive class (i.e. vaccine hesitancy) in each cross validation split (the standard deviations are mentioned between brackets). For the cross-genre experiments, on the other hand, results are reported on the test sets. In cases where CoNTACT outperformed RobBERT, p-values were calculated to determine whether the observed improvements are statistically significant \cite{mcnemar:1947}. 

As shown in the results, both models perform better on Twitter data than on Facebook data, and fine-tuning on both platforms simultaneously yields higher results than fine-tuning on the individual platforms. The standard deviations are, in spite of the small test sets, relatively small, which indicates consistent model performance. When comparing the results of RobBERT to those of CoNTACT, it can be observed that CoNTACT outperforms RobBERT in all experimental settings with statistical significance, including the cross-genre experiments. In other words, when fine-tuning on Twitter but testing on Facebook, CoNTACT strongly outperforms RobBERT, although no Facebook data was used during it's domain adaptation or fine-tuning phase. Additionally, CoNTACT outperforms RobBERT on Facebook data even if the former is fine-tuned on Twitter data and the latter is fine-tuned on Facebook data (i.e. data from the same platform). These results highlight the cross-genre potential of CoNTACT.

\begin{table}[t]
\centering
\caption{Results (\%) for vaccine hesitancy detection, including standard deviations (if applicable). The results are reported on the positive class, and statistically significant gains over the baseline are indicated with asterisks.}
\label{tab:stance_results}
\begin{tabular}{|c|l|l|l|l|l|l|}
\hline
\textbf{Model} & \textbf{Fine-tune} & \textbf{Test} & \textbf{Pre} & \textbf{Rec} & \textbf{F1} & \textbf{*} \\ \hline
\multirow{6}{*}{RobBERT} & Twitter & Twitter & 76.1 (3.6) & 74.2 (4.3) & 75.1 (3.1) & N/A \\
& Twitter & Facebook & 62.0 (-) & 59.8 (-) & 60.9 (-) & N/A \\
 & Facebook & Facebook & 69.5 (3.1) & 57.2 (3.2) & 62.7 (2.6) & N/A \\
 & Facebook & Twitter & 67.4 (-) & 63.0 (-) & 65.1 (-) & N/A \\
 & Both & Twitter & 77.1 (2.8) & 73.9 (4.0) & 75.4 (-) & N/A \\
 & Both & Facebook & 70.6 (3.5) & 64.6 (3.7) & 67.4 (2.7) & N/A \\ \hline
\multirow{6}{*}{CoNTACT} & Twitter & Twitter & 77.2 (3.5) & 76.9 (4.1) & 77.1 (3.6) & * \\ %0.037*
& Twitter & Facebook & 65.2 (-) & 64.9 (-) & 65.0 (-) & *** \\
 & Facebook & Facebook & 71.2 (3.2) & 67.5 (3.1) & 69.3 (2.9) & *** \\ %\textless{}0.001*
 & Facebook & Twitter & 71.0 (-) & 82.5 (-) & 76.3 (-) & ***\\
 & Both & Twitter & 78.9 (4.2) & 77.4 (1.7) & 78.1 (2.5) & ** \\ %0.005*
 & Both & Facebook & 73.2 (3.0) & 68.2 (4.3) & 70.6 (2.6) & ** \\ \hline %0.002*
\end{tabular}
\end{table}

In order to gain insight into which specific improvements CoNTACT made, a manual analysis\footnote{All examples provided below were translated from Dutch to English.} of the instances where CoNTACT classified vaccine stance correctly, and RobBERT did not, was conducted (for all experiments). False negatives, i.e., the cases where RobBERT did not predict vaccine hesitancy, but CoNTACT did (correctly), were the largest group of errors. They were found in vaccine-hesitant instances referring to pro-vaccination opinions, such as ``'We do not have evidence that vaccines cause damage to pregnant women so we advise pregnant women to get vaccinated', what kind of an idiot says things like this?!". Further, false negatives were caused by sarcasm and other forms of implicit language, e.g. ``they should start [the vaccination campaign] in Den Hague... double dosis". This message seems to express pro-vaccination opinions on a superficial level, but the author actually hopes that the government (located in Den Hague) will suffer from major side effects of the vaccine.

In comparison, false positives, i.e., cases where RobBERT incorrectly detected vaccine hesitancy, but CoNTACT correctly did not detect vaccine hesitancy, were found in messages containing certain hashtags or terms that are associated with vaccine hesitancy. For example, the tweet ``\#vaccinationobligation, because infecting others is not a fundamental right", expresses a pro-vaccination opinion. RobBERT, however, incorrectly detected vaccine hesitancy in this tweet, presumably because of the hashtag ``\#vaccinationobligation", which occurs frequently in vaccine-hesitant messages. Especially in the cross-genre experiments where the models were fine-tuned on Facebook and tested on Twitter, RobBERT was frequently confused by vaccine related hashtags, causing both false positives and negatives, whereas CoNTACT showed more understanding of said hashtags, even when both pro- and anti-vaccination hashtags appeared in the same message. Other false positives by the baseline were found in cases where vaccine-hesitant opinions were quoted or referred to, such as ``'poison vaccine', yeah right, you're so childish". Similarly, pro-vaccination messages expressing a negative sentiment towards, for example, vaccination policy, were misclassified more often by RobBERT than by CoNTACT, e.g. ``I am \#provaccination but I support protest against the mismanagement of the government".

An additional analysis of the comments where CoNTACT failed to correctly predict the stance but RobBERT did not was conducted. False negatives (the smallest group of errors) were found in messages using implicit or sarcastic language, such as ``this press conference was very clear as always...", similarly to the false negatives found in RobBERT. Regarding the false positives, the largest group of errors, we observed that there were cases where specific terms used frequently in vaccine-hesitant messages caused confusion, as was also observed in the error analysis of RobBERT. For example, in ``those \#SideEffects are not as bad as people think" and ``\#vaccineobligation is a must", CoNTACT interprets the hashtags as indicators for vaccine hesitancy, because it has learned this during the fine-tuning period. In conclusion, we observed that the models have difficulties with the same types of comments: messages containing forms implicit language caused false negative errors, whereas domain-specific terminology caused false positive errors. CoNTACT, however, made significantly less errors in these challenging cases due to the domain adaptation, indicating that CoNTACT has improved representations of COVID-19 related terminology.

\begin{table}[t]
\centering
\caption{Precision, recall, F1-score and EMR of RobBERT and CoNTACT for the vaccine argument experiments.}
\begin{tabular}{|c|l|l|l|l|l|l|}
\hline
\textbf{Model} & \textbf{Fine-tune} & \textbf{Test} & \textbf{Pre} & \textbf{Rec} & \textbf{F1} & \textbf{EMR}\\ \hline
\multirow{6}{*}{RobBERT} & Twitter & Twitter       & 62.5 (0.8)  & 50.2 (1.4) & 55.0 (1.0) & 46.7 (6.1) \\
              & Twitter & Facebook      & 50.7 (-) & 29.7 (-) & 36.3 (-) & 24.2 (-) \\
       & Facebook & Facebook     & 48.4 (1.4) & 31.7 (1.8) & 37.3 (1.7) & 34.9 (1.2) \\
              & Facebook & Twitter      & 59.5 (-)& 30.0 (-)& 33.3 (-)& 33.8 (-)\\
              & Both & Twitter          & 62.9 (1.5) & 53.4 (0.5) & 57.3 (0.8) & 47.7 (0.8) \\
              & Both & Facebook         & 56.6 (2.8) & 43.9 (3.1) & 48.9 (2.9) & 39.3 (1.6) \\ \hline
\multirow{6}{*}{CoNTACT} & Twitter & Twitter       & 64.7 (1.5) & 56.2 (0.9) & 59.8 (0.9) & 49.2 (1.3)\\  & Twitter & Facebook      & 56.9 (-) & 36.1 (-) & 42.7 (-)& 26.9 (-)\\
              & Facebook & Facebook     & 55.5 (5.9) & 41.1 (1.2) & 46.2 (1.9) & 41.0 (1.1) \\
              & Facebook & Twitter      & 57.5 (-)& 39.4 (-)& 41.4 (-)& 34.5 (-) \\
              & Both & Twitter          & 64.1 (1.3) & 58.4 (1.6) & 60.9 (0.9)  & 49.5 (1.1) \\
              & Both & Facebook         & 60.1 (3.3) & 49.7 (2.2) & 53.9 (2.4) & 41.9 (1.1)\\\hline
\end{tabular}
\label{tab:argument_results}
\end{table}

\subsection{Argument classification}
In Table \ref{tab:argument_results} the results on argument classification are summarised. Precision, recall, F1 (incl. standard deviations), and exact match ratio (EMR), an accuracy score for cases where the entire set of labels was predicted correctly, are reported. Overall, both models perform better on Twitter than on Facebook data, including the cross-genre experiments, similarly to the stance classification experiments. Although CoNTACT outperforms RobBERT in the cross-genre experiments, the results are still noticeably lower than the in-genre experiments. Further, fine-tuning on both Facebook and Twitter simultaneously increases model performance. When comparing the models, it can be observed that CoNTACT outperforms RobBERT in all experiments.

The results for the individual arguments for RobBERT and CoNTACT are presented in Table \ref{tab:results-arguments-robbert} and \ref{tab:results-arguments-contact}, respectively. The provided results are the results on the positive classes in the test set(s). Regarding the baseline results, it can be observed that certain classes are predicted substantially better than others. Overall, RobBERT predicted the ``safety" and ``liberty" classes best, whereas the most difficult classes were ``development" and ``alternative medicine" (these were also the most underrepresented classes in our data). 

When comparing the results of RobBERT to those of CoNTACT, an increase in performance on all classes in all experiments can be observed, except for ``conspiracy theory" in Twitter when fine-tuning on both platforms, ``alternative medicine" in Twitter when fine-tuning on Facebook, and ``institutional motives" in Facebook when fine-tuning on Twitter. Some of the highest improvements were found in the ``development" and ``alternative medicine" classes, which are the most challenging classes, as mentioned above. In order to verify whether the observed improvements are significant, a McNemar \cite{mcnemar:1947} test was conducted per argument class (Table \ref{tab:significance-arguments}). Despite the substantial gains, less than half of the improvements were considered statistically significant for the same- and mixed-genre experiments. We suspect that the significance test we used yielded higher p-values because the frequency of certain classes was too low to ascertain that improvements were significant rather than random. Further experiments with more data could therefore produce other results and new insights in the future. In the cross-genre experiments, however, CoNTACT showed statistically significant improvements on half of the argumentation classes when the model was fine-tuned on Twitter data and tested on Facebook data. Moreover, statistically significant improvements were observed for all classes when the model was fine-tuned on Facebook data and tested on Twitter data. These results highlight the cross-genre potential of CoNTACT.

\begin{table}[]
\small
\centering
\caption{Averaged results (\%) of RobBERT on each argument class per experiment.}
\begin{tabular}{l|ccc||ccc||ccc||ccc||ccc||ccc|}
\cline{2-19}
\textbf{} &
  \multicolumn{3}{c||}{\textbf{tw-tw}} &
  \multicolumn{3}{c||}{\textbf{tw-fb}} &
  \multicolumn{3}{c||}{\textbf{fb-fb}} &
  \multicolumn{3}{c||}{\textbf{fb-tw}} &
  \multicolumn{3}{c||}{\textbf{both-tw}} &
  \multicolumn{3}{c|}{\textbf{both-fb}} \\ \cline{2-19} 
 &
  \multicolumn{1}{c|}{\textbf{P}} &
  \multicolumn{1}{c|}{\textbf{R}} &
  \textbf{F} &
  \multicolumn{1}{c|}{\textbf{P}} &
  \multicolumn{1}{c|}{\textbf{R}} &
  \multicolumn{1}{c||}{\textbf{F}} &
  \multicolumn{1}{c|}{\textbf{P}} &
  \multicolumn{1}{c|}{\textbf{R}} &
  \textbf{F} &
  \multicolumn{1}{c|}{\textbf{P}} &
  \multicolumn{1}{c|}{\textbf{R}} &
  \multicolumn{1}{c||}{\textbf{F}} &
  \multicolumn{1}{c|}{\textbf{P}} &
  \multicolumn{1}{c|}{\textbf{R}} &
  \textbf{F} &
  \multicolumn{1}{c|}{\textbf{P}} &
  \multicolumn{1}{c|}{\textbf{R}} &
  \textbf{F} \\ \hline
\multicolumn{1}{|l|}{\textbf{alt.}} &
  \multicolumn{1}{c|}{60} &
  \multicolumn{1}{c|}{35} &
  45 &
  \multicolumn{1}{l|}{33} &
  \multicolumn{1}{l|}{36} &
  35 &
  \multicolumn{1}{c|}{0} &
  \multicolumn{1}{c|}{0} &
  0 &
  \multicolumn{1}{c|}{100} &
  \multicolumn{1}{c|}{6} &
  11 &
  \multicolumn{1}{c|}{66} &
  \multicolumn{1}{c|}{46} &
  54 &
  \multicolumn{1}{c|}{45} &
  \multicolumn{1}{c|}{32} &
  38 \\ \hline
\multicolumn{1}{|l|}{\textbf{con.}} &
  \multicolumn{1}{c|}{61} &
  \multicolumn{1}{c|}{43} &
  50 &
  \multicolumn{1}{c|}{32} &
  \multicolumn{1}{c|}{14} &
  19 &
  \multicolumn{1}{c|}{48} &
  \multicolumn{1}{c|}{22} &
  30 &
  \multicolumn{1}{c|}{34} &
  \multicolumn{1}{c|}{10} &
  15 &
  \multicolumn{1}{c|}{59} &
  \multicolumn{1}{c|}{46} &
  52 &
  \multicolumn{1}{c|}{54} &
  \multicolumn{1}{c|}{38} &
  45 \\ \hline
\multicolumn{1}{|l|}{\textbf{crit.}} &
  \multicolumn{1}{c|}{47} &
  \multicolumn{1}{c|}{27} &
  34 &
  \multicolumn{1}{c|}{45} &
  \multicolumn{1}{c|}{26} &
  33 &
  \multicolumn{1}{c|}{57} &
  \multicolumn{1}{c|}{49} &
  53 &
  \multicolumn{1}{c|}{27} &
  \multicolumn{1}{c|}{36} &
  31 &
  \multicolumn{1}{c|}{49} &
  \multicolumn{1}{c|}{30} &
  37 &
  \multicolumn{1}{c|}{59} &
  \multicolumn{1}{c|}{51} &
  55 \\ \hline
\multicolumn{1}{|l|}{\textbf{dev.}} &
  \multicolumn{1}{c|}{54} &
  \multicolumn{1}{c|}{38} &
  45 &
  \multicolumn{1}{c|}{42} &
  \multicolumn{1}{c|}{18} &
  25 &
  \multicolumn{1}{c|}{50} &
  \multicolumn{1}{c|}{18} &
  26 &
  \multicolumn{1}{c|}{54} &
  \multicolumn{1}{c|}{18} &
  27 &
  \multicolumn{1}{c|}{55} &
  \multicolumn{1}{c|}{45} &
  49 &
  \multicolumn{1}{c|}{47} &
  \multicolumn{1}{c|}{31} &
  37 \\ \hline
\multicolumn{1}{|l|}{\textbf{eff.}} &
  \multicolumn{1}{c|}{63} &
  \multicolumn{1}{c|}{54} &
  58 &
  \multicolumn{1}{c|}{50} &
  \multicolumn{1}{c|}{46} &
  48 &
  \multicolumn{1}{c|}{52} &
  \multicolumn{1}{c|}{36} &
  42 &
  \multicolumn{1}{c|}{61} &
  \multicolumn{1}{c|}{31} &
  42 &
  \multicolumn{1}{c|}{61} &
  \multicolumn{1}{c|}{55} &
  58 &
  \multicolumn{1}{c|}{58} &
  \multicolumn{1}{c|}{46} &
  51 \\ \hline
\multicolumn{1}{|l|}{\textbf{inst.}} &
  \multicolumn{1}{c|}{66} &
  \multicolumn{1}{c|}{60} &
  63 &
  \multicolumn{1}{c|}{59} &
  \multicolumn{1}{c|}{27} &
  37 &
  \multicolumn{1}{c|}{59} &
  \multicolumn{1}{c|}{32} &
  41 &
  \multicolumn{1}{c|}{76} &
  \multicolumn{1}{c|}{9} &
  17 &
  \multicolumn{1}{c|}{66} &
  \multicolumn{1}{c|}{62} &
  64 &
  \multicolumn{1}{c|}{60} &
  \multicolumn{1}{c|}{38} &
  46 \\ \hline
\multicolumn{1}{|l|}{\textbf{lib.}} &
  \multicolumn{1}{c|}{77} &
  \multicolumn{1}{c|}{76} &
  77 &
  \multicolumn{1}{c|}{61} &
  \multicolumn{1}{c|}{36} &
  46 &
  \multicolumn{1}{c|}{61} &
  \multicolumn{1}{c|}{48} &
54 &\multicolumn{1}{c|}{64} &\multicolumn{1}{c|}{83} & 72 &\multicolumn{1}{c|}{77} &\multicolumn{1}{c|}{78} & 77 &\multicolumn{1}{c|}{64} &\multicolumn{1}{c|}{49} &
  56 \\ \hline\multicolumn{1}{|l|}{\textbf{saf.}} &\multicolumn{1}{c|}{71} &\multicolumn{1}{c|}{67} & 69 &\multicolumn{1}{c|}{84} &\multicolumn{1}{c|}{34} & 49 &
  \multicolumn{1}{c|}{66} &\multicolumn{1}{c|}{63} & 64 &\multicolumn{1}{c|}{60} &\multicolumn{1}{c|}{47} & 53 &\multicolumn{1}{c|}{70} &\multicolumn{1}{c|}{67} & 69 &\multicolumn{1}{c|}{67} &
\multicolumn{1}{c|}{66} & 67 \\ \hline \hline\multicolumn{1}{|l|}{\textbf{micro}} &\multicolumn{1}{c|}{69} &\multicolumn{1}{c|}{60} & 64 &\multicolumn{1}{c|}{58} &\multicolumn{1}{c|}{30} & 39 &\multicolumn{1}{c|}{59} &\multicolumn{1}{c|}{44} & 50 &\multicolumn{1}{c|}{57} &\multicolumn{1}{c|}{45} & 50 &\multicolumn{1}{c|}{68} &\multicolumn{1}{c|}{62} & 65 &\multicolumn{1}{c|}{61} &\multicolumn{1}{c|}{51} & 56 \\ \hline\multicolumn{1}{|l|}{\textbf{macro}} &\multicolumn{1}{c|}{62} &\multicolumn{1}{c|}{50} & 55 &\multicolumn{1}{c|}{51} &\multicolumn{1}{c|}{30} & 36 &\multicolumn{1}{c|}{48} &\multicolumn{1}{c|}{32} & 37 &\multicolumn{1}{c|}{60} &\multicolumn{1}{c|}{30} & 33 &\multicolumn{1}{c|}{63} &\multicolumn{1}{c|}{53} & 57 &\multicolumn{1}{c|}{60} &\multicolumn{1}{c|}{44} & 49 \\ \hline\multicolumn{1}{|l|}{\textbf{weighted}} &\multicolumn{1}{c|}{67} &\multicolumn{1}{c|}{60} & 63 &\multicolumn{1}{c|}{61} &\multicolumn{1}{c|}{30} & 39 &\multicolumn{1}{c|}{57} &\multicolumn{1}{c|}{44} & 48 &\multicolumn{1}{c|}{60} &\multicolumn{1}{c|}{45} & 45 &\multicolumn{1}{c|}{67} &\multicolumn{1}{c|}{62} & 64 &\multicolumn{1}{c|}{60} &\multicolumn{1}{c|}{51} & 55 \\ \hline\multicolumn{1}{|l|}{\textbf{samples}} &\multicolumn{1}{c|}{56} &\multicolumn{1}{c|}{53} & 53 &\multicolumn{1}{c|}{32} &\multicolumn{1}{c|}{28} & 29 &\multicolumn{1}{c|}{50} &\multicolumn{1}{c|}{45} & 46 &\multicolumn{1}{c|}{50} &\multicolumn{1}{c|}{42} & 44 &
  \multicolumn{1}{c|}{58} &\multicolumn{1}{c|}{55} & 55 &\multicolumn{1}{c|}{52} &\multicolumn{1}{c|}{49} & 49 \\ \hline
\end{tabular}
\label{tab:results-arguments-robbert}
\end{table}

\begin{table}[]
\centering
\caption{Averaged results (\%) of CoNTACT on each argument class per experiment.}
\begin{tabular}{l|ccc||ccc||ccc||ccc||ccc||ccc|}
\cline{2-19}
\textbf{} & \multicolumn{3}{c||}{\textbf{tw-tw}} & \multicolumn{3}{c||}{\textbf{tw-fb}} & \multicolumn{3}{c||}{\textbf{fb-fb}} & \multicolumn{3}{c||}{\textbf{fb-tw}} & \multicolumn{3}{c||}{\textbf{both-tw}} & \multicolumn{3}{c|}{\textbf{both-fb}} \\ \cline{2-19} 
 & \multicolumn{1}{c|}{\textbf{P}} & \multicolumn{1}{c|}{\textbf{R}} & \textbf{F} & \multicolumn{1}{c|}{\textbf{P}} & \multicolumn{1}{c|}{\textbf{R}} & \multicolumn{1}{c||}{\textbf{F}} & \multicolumn{1}{c|}{\textbf{P}} & \multicolumn{1}{c|}{\textbf{R}} & \textbf{F} & \multicolumn{1}{c|}{\textbf{P}} & \multicolumn{1}{c|}{\textbf{R}} & \multicolumn{1}{c||}{\textbf{F}} & \multicolumn{1}{c|}{\textbf{P}} & \multicolumn{1}{c|}{\textbf{R}} & \textbf{F} & \multicolumn{1}{c|}{\textbf{P}} & \multicolumn{1}{c|}{\textbf{R}} & \textbf{F} \\ \hline
\multicolumn{1}{|l|}{\textbf{alt.}} & \multicolumn{1}{l|}{67} & \multicolumn{1}{l|}{48} & 56 & \multicolumn{1}{l|}{42} & \multicolumn{1}{l|}{46} & 44 & \multicolumn{1}{l|}{67} & \multicolumn{1}{c|}{4} & 7 & \multicolumn{1}{l|}{50} & \multicolumn{1}{l|}{3} & 5 & \multicolumn{1}{l|}{64} & \multicolumn{1}{l|}{56} & 60 & \multicolumn{1}{l|}{55} & \multicolumn{1}{l|}{46} & 51 \\ \hline
\multicolumn{1}{|l|}{\textbf{con.}} & \multicolumn{1}{l|}{60} & \multicolumn{1}{l|}{49} & 54 & \multicolumn{1}{l|}{32} & \multicolumn{1}{l|}{23} & 27 & \multicolumn{1}{l|}{55} & \multicolumn{1}{l|}{31} & 40 & \multicolumn{1}{l|}{52} & \multicolumn{1}{l|}{31} & 39 & \multicolumn{1}{l|}{57} & \multicolumn{1}{l|}{47} & 51 & \multicolumn{1}{l|}{54} & \multicolumn{1}{l|}{43} & 48 \\ \hline
\multicolumn{1}{|l|}{\textbf{crit.}} & \multicolumn{1}{l|}{51} & \multicolumn{1}{l|}{34} & 41 & \multicolumn{1}{l|}{49} & \multicolumn{1}{l|}{42} & 45 & \multicolumn{1}{l|}{61} & \multicolumn{1}{l|}{57} & 59 & \multicolumn{1}{l|}{25} & \multicolumn{1}{l|}{46} & 33 & \multicolumn{1}{l|}{49} & \multicolumn{1}{l|}{37} & 42 & \multicolumn{1}{l|}{59} & \multicolumn{1}{l|}{58} & 59 \\ \hline
\multicolumn{1}{|l|}{\textbf{dev.}} & \multicolumn{1}{l|}{58} & \multicolumn{1}{l|}{47} & 52 & \multicolumn{1}{l|}{56} & \multicolumn{1}{l|}{24} & 36 & \multicolumn{1}{l|}{55} & \multicolumn{1}{l|}{33} & 41 & \multicolumn{1}{l|}{56} & \multicolumn{1}{l|}{36} & 44 & \multicolumn{1}{l|}{58} & \multicolumn{1}{l|}{51} & 54 & \multicolumn{1}{l|}{54} & \multicolumn{1}{l|}{34} & 42 \\ \hline
\multicolumn{1}{|l|}{\textbf{eff}} & \multicolumn{1}{l|}{64} & \multicolumn{1}{l|}{62} & 63 & \multicolumn{1}{l|}{61} & \multicolumn{1}{l|}{57} & 59 & \multicolumn{1}{l|}{61} & \multicolumn{1}{l|}{50} & 55 & \multicolumn{1}{l|}{69} & \multicolumn{1}{l|}{50} & 58 & \multicolumn{1}{l|}{65} & \multicolumn{1}{l|}{65} & 65 & \multicolumn{1}{l|}{59} & \multicolumn{1}{l|}{53} & 56 \\ \hline
\multicolumn{1}{|l|}{\textbf{inst.}} & \multicolumn{1}{l|}{68} & \multicolumn{1}{l|}{63} & 66 & \multicolumn{1}{l|}{61} & \multicolumn{1}{l|}{22} & 32 & \multicolumn{1}{l|}{57} & \multicolumn{1}{l|}{38} & 46 & \multicolumn{1}{l|}{78} & \multicolumn{1}{l|}{10} & 18 & \multicolumn{1}{l|}{68} & \multicolumn{1}{l|}{63} & 66 & \multicolumn{1}{l|}{62} & \multicolumn{1}{l|}{42} & 50 \\ \hline
\multicolumn{1}{|l|}{\textbf{lib.}} & \multicolumn{1}{l|}{78} & \multicolumn{1}{l|}{77} & 78 & \multicolumn{1}{l|}{67} & \multicolumn{1}{l|}{38} & 49 & \multicolumn{1}{l|}{66} & \multicolumn{1}{l|}{50} & 57 & \multicolumn{1}{l|}{72} & \multicolumn{1}{l|}{81} & 76 & \multicolumn{1}{l|}{78} & \multicolumn{1}{l|}{78} & 78 & \multicolumn{1}{l|}{65} & \multicolumn{1}{l|}{51} & 57 \\ \hline
\multicolumn{1}{|l|}{\textbf{saf.}} & \multicolumn{1}{l|}{72} & \multicolumn{1}{l|}{69} & 71 & \multicolumn{1}{l|}{87} & \multicolumn{1}{l|}{37} & 52 & \multicolumn{1}{l|}{70} & \multicolumn{1}{l|}{67} & 69 & \multicolumn{1}{l|}{58} & \multicolumn{1}{l|}{57} & 58 & \multicolumn{1}{l|}{72} & \multicolumn{1}{l|}{71} & 72 & \multicolumn{1}{l|}{70} & \multicolumn{1}{l|}{70} & 70 \\ \hline \hline
\multicolumn{1}{|l|}{\textbf{micro}} & \multicolumn{1}{l|}{70} & \multicolumn{1}{l|}{64} & 67 & \multicolumn{1}{l|}{62} & \multicolumn{1}{l|}{36} & 46 & \multicolumn{1}{l|}{64} & \multicolumn{1}{l|}{53} & 58 & \multicolumn{1}{l|}{58} & \multicolumn{1}{l|}{51} & 55 & \multicolumn{1}{l|}{69} & \multicolumn{1}{l|}{65} & 67 & \multicolumn{1}{l|}{63} & \multicolumn{1}{l|}{56} & 59 \\ \hline
\multicolumn{1}{|l|}{\textbf{macro}} & \multicolumn{1}{l|}{65} & \multicolumn{1}{l|}{56} & 60 & \multicolumn{1}{l|}{57} & \multicolumn{1}{l|}{36} & 43 & \multicolumn{1}{l|}{61} & \multicolumn{1}{l|}{41} & 47 & \multicolumn{1}{l|}{58} & \multicolumn{1}{l|}{39} & 41 & \multicolumn{1}{l|}{64} & \multicolumn{1}{l|}{58} & 61 & \multicolumn{1}{l|}{60} & \multicolumn{1}{l|}{50} & 54 \\ \hline
\multicolumn{1}{|l|}{\textbf{weighted}} & \multicolumn{1}{l|}{69} & \multicolumn{1}{l|}{64} & 66 & \multicolumn{1}{l|}{66} & \multicolumn{1}{l|}{36} & 46 & \multicolumn{1}{l|}{63} & \multicolumn{1}{l|}{53} & 57 & \multicolumn{1}{l|}{63} & \multicolumn{1}{l|}{51} & 51 & \multicolumn{1}{l|}{69} & \multicolumn{1}{l|}{65} & 67 & \multicolumn{1}{l|}{63} & \multicolumn{1}{l|}{56} & 59 \\ \hline
\multicolumn{1}{|l|}{\textbf{samples}} & \multicolumn{1}{l|}{58} & \multicolumn{1}{l|}{57} & 56 & \multicolumn{1}{l|}{37} & \multicolumn{1}{l|}{34} & 34 & \multicolumn{1}{l|}{55} & \multicolumn{1}{l|}{51} & 51 & \multicolumn{1}{l|}{53} & \multicolumn{1}{l|}{48} & 48 & \multicolumn{1}{l|}{59} & \multicolumn{1}{l|}{58} & 57 & \multicolumn{1}{l|}{56} & \multicolumn{1}{l|}{54} & 53 \\ \hline
\end{tabular}
\label{tab:results-arguments-contact}
\end{table}

In order to gain insight into the specific improvements of CoNTACT, a manual error analysis of the predictions of both models was conducted. First, instances where CoNTACT succeeded and RobBERT failed to predict the correct argument(s) were investigated. For each argument class, several terms seemed to guide the predictions of CoNTACT, because of the learned representations of said terms during both the domain adaptation and fine-tuning phase. For instance, references to the immune system and drugs, such as Ivermectine, were found to be stronger indicators of the ``alternative medicine" argument class for CoNTACT than for RobBERT in predicting this argument. Further, comments containing words and hashtags such as ``medical experiment" and ``lab rat" were classified correctly by CoNTACT as related to ``development", contrary to RobBERT, which made more false negative errors in this class. Similar observations were made for ``institutional motives" (e.g. references to governments, political parties and politicians, such as \#rutte3, \#dv66 and \#hugodejonge), ``conspiracy theory" (e.g. references to gene therapy, such as ``\#geneticmodification"),  ``safety" (also references to gene therapy), and ``liberty" (e.g. references to vaccine passports and obligation). 

In addition, messages where RobBERT predicted the correct arguments but CoNTACT did not were investigated, although no clear error patterns were found in these cases. In general, however, both models seem to incorrectly classify arguments when the message itself lacks context or terminology related to the argument. For example, in the Facebook comment ``they don't want them [the vaccines] anywhere else", which was annotated with the ``criticism on vaccination strategy" label, both models failed to predict any argument, since the reference to e.g., a potential surplus of vaccines is implicit in this case.

In conclusion, CoNTACT seems to have learned domain-specific terminology in the domain adaptation phase, which benefits the model for the argument detection task, as can be derived from the results. The error analysis, however, showed that the model still experiences difficulties with classifying text entries that lack context or explicit information about the relevant argument(s).

\begin{table}[]
\centering
\caption{Statistically significant improvements in the argumentation detection task of CoNTACT over RobBERT.}
\begin{tabular}{|l|l|}
\hline
\textbf{Experiment} & \textbf{Classes with significant improvements} \\ \hline
\textbf{Tw - Tw} & efficacy (***) \\ \hline
\textbf{Tw - Fb} & \begin{tabular}[c]{@{}l@{}}conspiracy (*), criticism on vaccination strategy (***),\\ institutional motives (***), liberty (***)\end{tabular} \\ \hline
\textbf{Fb - Fb} & \begin{tabular}[c]{@{}l@{}}development (***), efficacy (***), institutional motives (***), \\ liberty (***), safety (*)\end{tabular} \\ \hline
\textbf{Fb - Tw} & \begin{tabular}[c]{@{}l@{}}alternative medicine (***), conspiracy (***), \\ criticism on vaccination strategy (***), development (***), \\ efficacy (***), institutional motives (***), liberty (***), safety (*)\end{tabular} \\ \hline
\textbf{Both - Tw} & efficacy (**) \\ \hline
\textbf{Both - Fb} & \begin{tabular}[c]{@{}l@{}}criticism on vaccination strategy (***), development (*)\end{tabular} \\ \hline
\end{tabular}
\label{tab:significance-arguments}
\end{table}

\clearpage
\section{Conclusion}
In this work we presented CoNTACT, a Dutch language model adapted to the domain of COVID-19 tweets. The model was developed by continuing the masked language modeling pre-training phase of RobBERT using 2.8M Dutch tweets related to COVID-19. In order to test the performance of CoNTACT, the model was tested on two classification tasks: detection of vaccine hesitancy and detection of arguments for vaccine hesitancy. These tasked were performed in various experimental settings, that is by fine-tuning and testing on social media messages from two different platforms: Twitter and Facebook. For the vaccine hesitancy detection task, CoNTACT outperformed RobBERT with statistical significance in all experiments, including cross-genre settings. With respect to the argument classification task, CoNTACT showed substantial gains in virtually all classes in all experiments, some of which with statistical significance. An error analysis showed that the domain adaptation resulted in better representations of COVID-19 related terminology, and therefore in better results. Issues remain in messages containing implicit/figurative language or messages lacking context. Future work may include the development of a second version of CoNTACT, where the model is fine-tuned on more data from various platforms (Twitter, Facebook, Reddit, etc.) for even more cross-genre robustness.

\bibliographystyle{splncs04}
\bibliography{mybibliography}
\end{document}